\documentclass[conference]{IEEEtran}
\IEEEoverridecommandlockouts

\usepackage{cite}
\usepackage{threeparttable}
\usepackage{microtype}
\usepackage{graphicx}
\usepackage{subfigure}
\usepackage{booktabs} 
\usepackage{makecell}
\usepackage{amsmath}
\usepackage{algorithm}
\usepackage{algpseudocode}
\usepackage{algorithmicx}
\usepackage{hyperref}
\usepackage{amsmath}
\usepackage{amsfonts}
\usepackage{amssymb}
\usepackage{amsmath,amssymb,amsfonts}
\usepackage{textcomp}
\usepackage{xcolor}
\usepackage{makecell}
\usepackage{tabularx}
\usepackage{textcomp}
\usepackage{xcolor}
\usepackage{fancyhdr}
\usepackage{marvosym}
\usepackage{multirow}

\def\BibTeX{{\rm B\kern-.05em{\sc i\kern-.025em b}\kern-.08em
    T\kern-.1667em\lower.7ex\hbox{E}\kern-.125emX}}
\begin{document}

\title{MoBiLE: Efficient Mixture-of-Experts Inference on Consumer GPU with \underline{M}ixture \underline{o}f \underline{Bi}g \underline{L}ittle \underline{E}xperts}

\author{\IEEEauthorblockN{Yushu Zhao$^{*}$, Yubin Qin$^{*}$, Yang Wang\textsuperscript{\Letter}, Xiaolong Yang, Huiming Han, Shaojun Wei, Yang Hu, Shouyi Yin\textsuperscript{\Letter}}
\IEEEauthorblockA{BNRist, Tsinghua University}
\IEEEauthorblockA{Email: wangyang\_imec@mail.tsinghua.edu.cn, yinsy@tsinghua.edu.cn}
\thanks{* Equal contribution.}

}

\maketitle

\begin{abstract}
Mixture-of-Experts (MoE) models have recently demonstrated exceptional performance across a diverse range of applications. The principle of sparse activation in MoE models facilitates an offloading strategy, wherein active experts are maintained in GPU HBM, while inactive experts are stored in CPU DRAM. The efficacy of this approach, however, is fundamentally constrained by the limited bandwidth of the CPU-GPU interconnect.
To mitigate this bottleneck, existing approaches have employed prefetching to accelerate MoE inference. These methods attempt to predict and prefetch the required experts using specially trained modules. Nevertheless, such techniques are often encumbered by significant training overhead and have shown diminished effectiveness on recent MoE models with fine-grained expert segmentation. 

In this paper, we propose MoBiLE, a plug-and-play offloading-based MoE inference framework with \textit{mixture of big-little experts}. It reduces the number of experts for unimportant tokens to half for acceleration while maintaining full experts for important tokens to guarantee model quality. Further, a dedicated fallback and prefetching mechanism is designed for switching between little and big experts to improve memory efficiency.
We evaluate MoBiLE on four typical modern MoE architectures and challenging generative tasks. Our results show that MoBiLE achieves a speedup of \(1.60\times\) to \(1.72\times\)  compared to the baseline on a consumer GPU system, with negligible degradation in accuracy.

\end{abstract}
\begin{IEEEkeywords}
large language model, mixture-of-experts, offloading, algorithm-system co-design.
\end{IEEEkeywords}

\section{Introduction}
\begin{figure}[htbp]
    \centering
    \includegraphics[width=0.75\columnwidth,keepaspectratio]{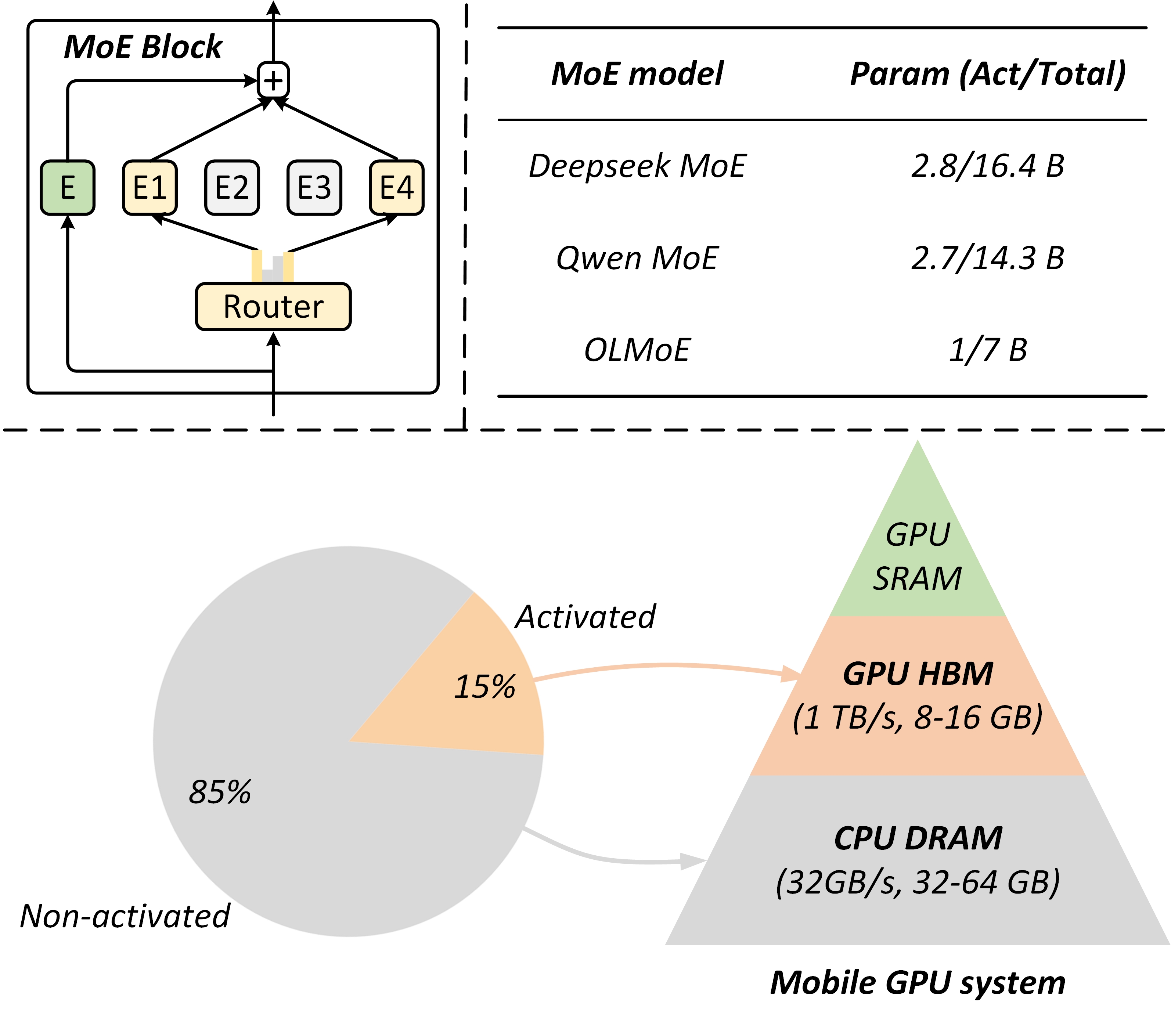} 
    \caption{Upper left: Illustration of the MoE module. The router dynamically determines which experts to activate based on the current input. 
    Upper right: parameter statistics of modern SOTA MoE models.
    Lower: Offloading-based inference for MoE models, where non-activated experts are stored in CPU DRAM, and the activated parameters are transferred to the GPU HBM for computation.}
    \label{fig1}
\end{figure}

Large language models (LLMs) have established state-of-the-art performance on a diverse array of natural language processing (NLP) tasks \cite{DBLP:journals/corr/abs-2407-06204}. This superior performance is primarily attributed to their substantial parameter sizes. 
To address the large computation challenge for LLMs, mixture-of-experts (MoE) models have been proposed. As illustrated in Fig. \ref{fig1}, MoE models enable large parameter size while keeping small computation effort by using a modular structure with multiple ``experts'' for the feed-forward networks (FFNs) in the model. During inference, a router dynamically selects a sparse subset of these experts for each input token. Such mechanism significantly reduces the computational burden during inference, thereby alleviating resource constraints.

Since only a small subset of the total parameters is needed for each token in MoE, it seems a good match to run MoE on consumer-level hardware, which typically has a GPU with HBM (8-16GB) and a CPU with relatively larger DRAM (32-64GB). For this purpose, researchers propose \textit{offloading} \cite{DBLP:conf/isca/HwangWCHTCY24, DBLP:conf/mlsys/DuLWJ0ZW00024, DBLP:journals/corr/abs-2408-10284}, which keeps activated parameters on faster GPU HBM, while storing the inactive ones in the slower CPU DRAM. These inactive parameters are then loaded into the GPU only when needed for computation.
As illustrated in Fig. \ref{fig1}, this hardware paradigm presents an opportunity to load only the active expert parameters onto the GPU for computation. For example, the Qwen MoE model \cite{qwen_moe} contains 14.3 billion parameters in total, yet only activates 2.7 billion for inference. This smaller, active subset is well within the storage capacity of typical GPU HBM. 
Consequently, a strategy of offloading inactive parameters to CPU DRAM and dynamically fetching them to the GPU as needed emerges as a promising approach for efficient MoE inference on consumer-level hardware systems. 

In offloading-based MoE inference, the dynamic fetching of experts from CPU DRAM to GPU HBM imposes a substantial performance bottleneck due to the limited bandwidth of the CPU-GPU interconnect. As documented in \cite{DBLP:journals/corr/abs-2411-01433}, the latency incurred by loading experts can constitute over 80\% of the total end-to-end latency. To address this challenge, prior works \cite{DBLP:conf/isca/HwangWCHTCY24, DBLP:conf/mlsys/DuLWJ0ZW00024, DBLP:journals/corr/abs-2308-14352} have proposed expert prefetching methodologies. These approaches utilize trained auxiliary modules to predict which experts will be activated in subsequent layers, enabling their preloading in advance. However, this strategy has two non-neglectable limitations. First, it requires specialized training to build the prediction module for each specific task, which is far from generality and incurs large computational overhead. Second, contemporary MoE architectures \cite{qwen_moe, DBLP:conf/acl/DaiDZXGCLZYWXLH24, DBLP:journals/corr/abs-2409-02060, DBLP:conf/icml/LudziejewskiKAP24} feature more experts as well as activated ones, which is not compatible with previous predicting methods and suffers from accuracy degradation \cite{DBLP:journals/corr/abs-2408-10284}. Consequently, existing MoE prefetching techniques are not only suboptimal for modern MoE models but also lack generality.

\begin{figure}[htbp]
    \centering
    \includegraphics[width=0.8\columnwidth,keepaspectratio]{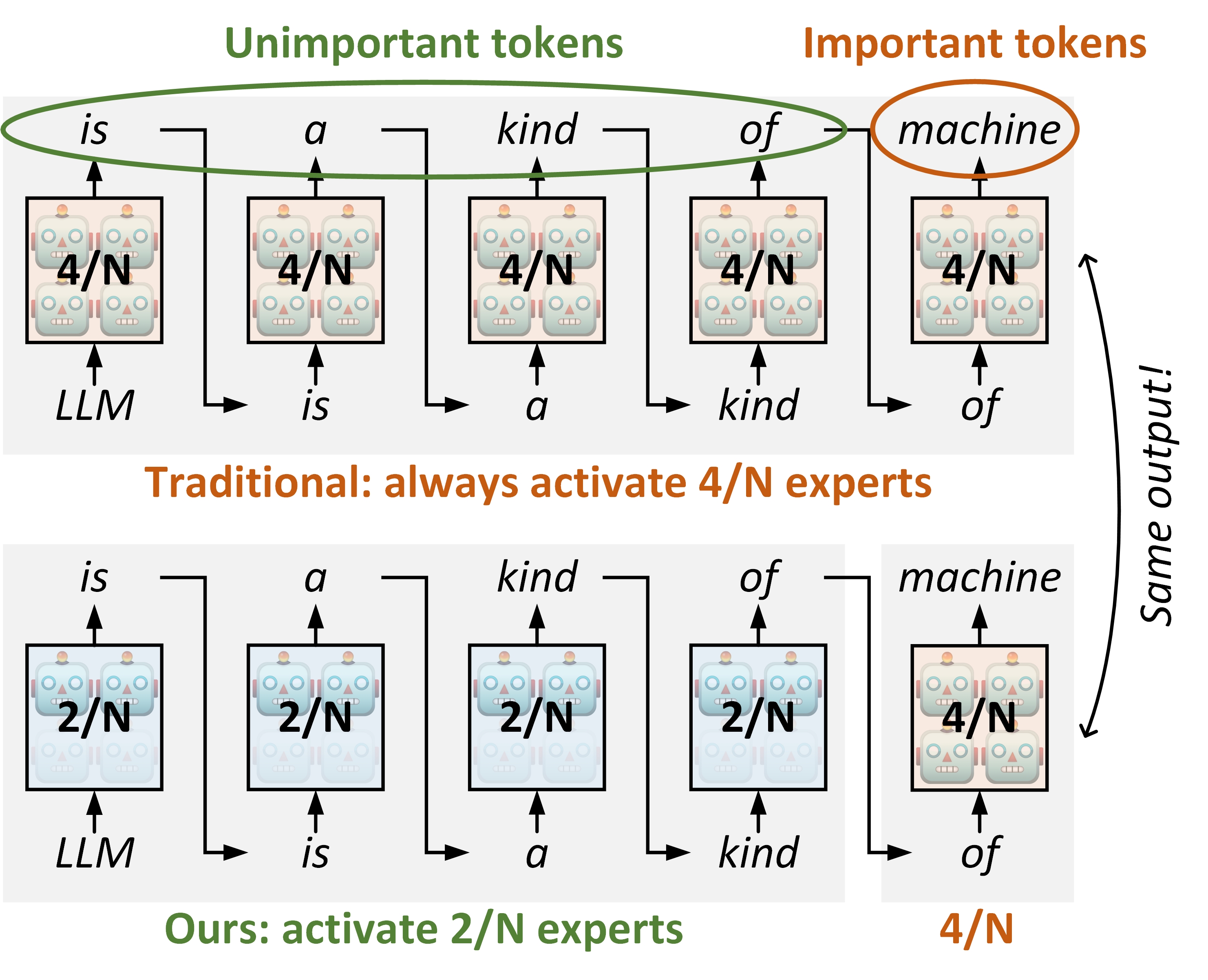} 
    \caption{For MoE models, activating less than the default number of activated experts leads to no difference for most of the tokens (unimportant tokens), while a small amount of important tokens need more experts to process.}
    \label{fig2}
\end{figure}

In this work, we propose MoBiLE, a plug-and-play framework designed to accelerate offloading-based MoE inference on consumer hardware. 
Our methodology is predicated on the observation that standard MoE models allocate a fixed number of experts to process every token, regardless of each token's informational significance. We posit that it is more efficient to assign a greater number of experts to salient tokens while utilizing fewer experts for those that are less critical, as shown in Fig. \ref{fig2}. 
Building on this insight, MoBiLE implements a ``big-little" expert allocation strategy. This approach reduces the number of activated experts for the majority of tokens to enhance efficiency. For a subset of critical tokens, it utilizes a fallback mechanism that dedicates the default number of activated experts to ensure high-fidelity processing. To mitigate the latency associated with this fallback, MoBiLE incorporates an efficient, training-free prefetching mechanism that overlaps computation with memory I/O, thereby accelerating the fallback path for these important tokens.

The main contributions of this paper are as follows:
\begin{itemize}
    \item We propose MoBiLE, a MoE inference system based on offloading technique, which utilizes a scheme of activating fewer experts for most tokens and performing fallback for the important tokens to gain better accuracy performance. To the best of our knowledge, MoBiLE is the first MoE inference system targeting consumer GPU supporting up-to-date MoE architecture.
    \item We further employ a training-free prefetching mechanism for the fallback process to overlap the memory and computation latency, utilizing the router logits from the discarded tokens. This approach enhances both accuracy and hardware efficiency.
    \item We demonstrate the efficacy of MoBiLE with state-of-the-art MoE models on consumer hardware systems, which shows a maximum \(1.72\times\) speedup while preserving the accuracy performance of the original model on generation tasks.
\end{itemize}

\section{Background and Motivation}
\label{Background and Motivation}

\subsection{MoE Architecture}
The MoE architecture provides substantial performance enhancements over conventional dense model designs.  As shown in Fig. \ref{fig1}, in a MoE model's decoder layer, instead of utilizing a dense MLP with a single feedforward network (FFN), the layer comprises multiple FFNs, referred to as ``experts". A router module dynamically determines which experts to activate based on the current input hidden states, and each expert is capable of processing distinct patterns of the input tokens. 
The execution of a MoE module occurs in two stages: the selection stage, where the router selects the necessary experts and produces routing weights, and the execution stage, during which the hidden states are passed through all selected experts and finally summed.
MoE achieves superior performance compared with dense models with similar activated parameters, primarily due to its ability to scale by adding expert parameters and increasing the total parameter size.

Recent advancements have marked a significant evolution in the design of MoE architectures. Traditionally, the number of activated experts in MoE models is relatively small. For instance, Google's Switch transformer \cite{DBLP:journals/jmlr/FedusZS22} activates only a single expert, and  Meta's NLLB-MoE \cite{DBLP:journals/corr/abs-2207-04672} activates 2 experts. 
However, a new trend has emerged in more recent open-source MoE models  \cite{qwen_moe, DBLP:conf/acl/DaiDZXGCLZYWXLH24, DBLP:journals/corr/abs-2409-02060}. The research behind these models indicates that employing a more fine-grained segmentation of experts enables each expert to acquire more specialized and distinct knowledge. This increased specialization has been shown to directly correlate with improved model performance, shifting the paradigm towards architectures with a larger number of more granular experts. 

\subsection{Large MoE Overhead and MoE Offloading Methods}
Although MoE models feature sparse activation and consequently lower computational requirements per inference, their overall parameter count remains substantial. This large model size makes it infeasible to store the entire model within the memory of a single consumer-grade GPU. 

To address the conflict between large model sizes and limited GPU memory, a promising strategy is to offload non-activated parameters to CPU memory and dynamically load the activated parameters to the GPU for computation. This approach is particularly viable given the typical memory asymmetry in consumer hardware: a GPU may offer 8-16 GB of high-bandwidth memory, whereas the system's CPU DRAM can readily provide 64 GB or more. 
By storing non-activated parameters in the CPU DRAM and loading the required experts at runtime, substantial GPU memory savings can be realized, thereby enabling MoE inference on consumer hardware devices.
As demonstrated in Fig. \ref{fig1}, for Deepseek MoE, OLMoE, and Qwen1.5 MoE, despite their substantial total parameter counts, the activated parameters remain relatively small, making them suitable for computation on consumer GPUs.

When utilizing an offloading-based approach for MoE inference, a significant performance bottleneck emerges due to the routing mechanism. In a MoE layer, the router dynamically determines which experts are required for the current input. These selected experts must then be fetched from CPU memory to the GPU's HBM at runtime.  This on-demand data transfer is severely constrained by the limited bandwidth of the CPU-GPU interconnect, such as the PCIe 4.0 interface, which offers a theoretical maximum bandwidth of approximately 64 GB/s. \cite{DBLP:journals/corr/abs-2411-01433} indicates that the time spent on loading experts can account for more than \(80\%\) of the total MoE module latency, representing a substantial overhead. 

Existing works address this challenge with various approaches.
\begin{itemize}
    \item \textbf{Pre-gated MoE \cite{DBLP:conf/isca/HwangWCHTCY24}} trains additional gating modules to predict the experts needed for the next MoE block and fetch the experts ahead. However, for the new MoE architectures with more experts activated, prefetching before a single layer may not totally overlap the experts' loading time, and for complicated tasks, prefetching the wrong experts may lead to a severe accuracy decrease. 
    \item \textbf{SIDA-MOE \cite{DBLP:conf/mlsys/DuLWJ0ZW00024}} proposes a data-aware algorithm to prefetch the experts needed for executing the next token. It trains an additional network containing LSTM and fully connected layers to build the hash function for expert prefetching. However, the incorrect prediction can lead to a large degradation in accuracy. 
    \item \textbf{AdapMoE \cite{DBLP:journals/corr/abs-2408-10284}} proposes an adaptive and sensitivity-based expert gating and prefetching method. However, to gain speedup, it uses a large cache on GPU to store the experts that may be needed, which leads to a large GPU memory overhead, causing low GPU utilization. For consumer GPUs, whose memory size is only slightly larger than the activated parameters, using a large cache to store the experts is impractical.
\end{itemize}

Based on the analysis of existing MoE offloading systems, we identify that a critical aspect of MoE offloading acceleration is the use of a plug-and-play method that does not require additional training and imposes minimal GPU memory overhead. This combination of features is essential to ensure practical deployment on the memory-constrained GPUs typically found in consumer-level devices.

\begin{figure*}[htbp]
    \centering
    \includegraphics[width=0.9\textwidth,keepaspectratio]{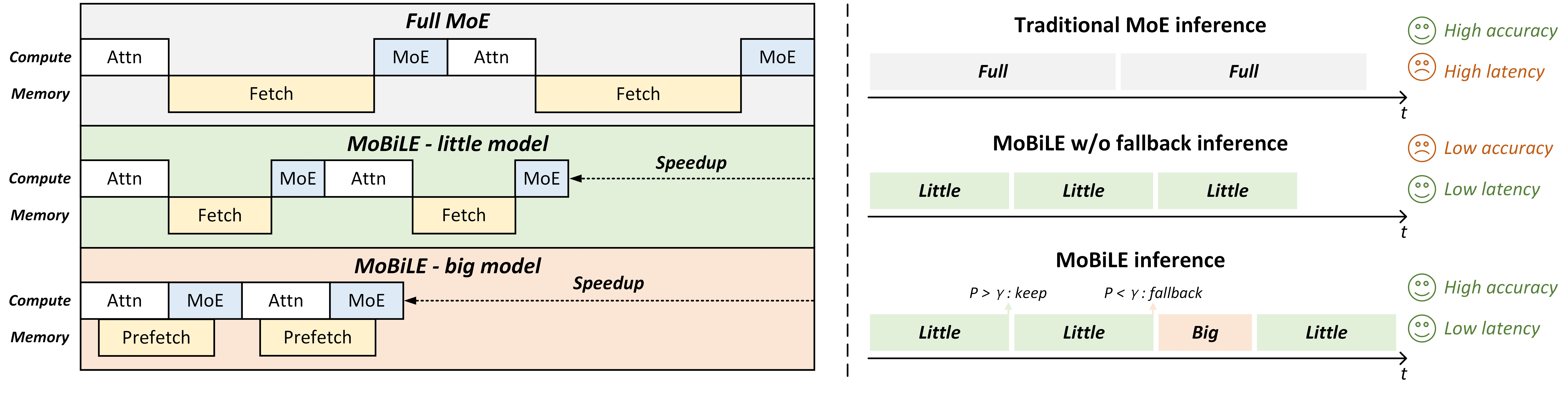} 
    \caption{Left: Latency comparison of different methods, including full MoE, MoBiLE little expert, and MoBiLE big expert. Right: comparison of different inference schemes. Traditional MoE inference suffers from high latency, while naively reducing the number of activated experts leads to a high accuracy decrease. MoBiLE maintains high accuracy while achieving low latency.}
    \label{fig3}
\end{figure*}

\subsection{Motivation}
MoE models employ a uniform number of experts for processing each token, regardless of the inherent variability in the information content of different tokens. This uniformity overlooks the fact that certain tokens play a more pivotal role in influencing overall model performance, while others may provide less critical information. 
As such, it stands to reason that allocating a greater number of experts to the more significant tokens could enhance processing efficacy, while the less important tokens could be adequately managed with fewer experts, as shown in Fig.\ref{fig2}. This differential allocation strategy has the potential to preserve model accuracy while significantly improving computational efficiency. 

This observation aligns with speculative decoding. In a speculative decoding framework of BiLD \cite{DBLP:conf/nips/KimMMMMGK23}, a smaller, efficient ``little" model generates the majority of tokens autoregressively. A larger, more powerful model then selectively reviews and regenerates only the most critical tokens.  This design ensures that the large model is only activated for key tokens, allowing the little model to effectively handle less critical tokens.
Recognizing this potential for improvement, we propose MoBiLE, an innovative offloading-based MoE inference system specifically tailored for consumer hardware. A detailed comparison of  MoBiLE and existing methods is shown in Table. \ref{table2}.

\begin{table}[t]
\begin{center}
\caption{Comparison of MoE Offloading Acceleration Methods.}
\label{table2}
\begin{tabular}{lccc}
\toprule
\textbf{Method} & \textbf{Requires train?} & \textbf{Models} & \textbf{Memory}  \\
\midrule
Pre-gated MoE & \textcolor{red}{Yes} & \textcolor{red}{Switch} & \textcolor{green}{Low} \\
Sida MoE & \textcolor{red}{Yes} & \textcolor{red}{Switch} & \textcolor{green}{Low} \\
AdapMoE & \textcolor{red}{Yes} & \textcolor{green}{Mixtral} & \textcolor{red}{High} \\
\textbf{MoBiLE (Ours)} & \textcolor{green}{No} & \textcolor{green}{Qwen MoE, OLMoE} & \textcolor{green}{Low}\\
\bottomrule
\end{tabular}

\end{center}
\end{table}

\section{MoBiLE Design} \label{MoBiLE Design}

In response to the challenges and motivations outlined above, we introduce our design, the Mixture of Big Little Experts (MoBiLE). MoBiLE builds upon the sparsity inherent in MoE modules by incorporating a fallback mechanism with training-free prefetching to optimize the performance of sparser MoE configurations. 
Specifically, MoBiLE dynamically transfers activated experts to the GPU, while offloading non-activated expert parameters to the CPU. This architecture facilitates efficient MoE inference on consumer-level GPU with limited HBM size, effectively addressing the memory constraints imposed by such hardware limitations.

\subsection{Overall Design}
Drawing inspiration from the principles of speculative decoding, we recognize that speculative decoding frameworks can be effectively applied to MoE models. Suppose the default number of activated experts is \(K\), in this context, the model activating \(K\) experts can be regarded as the ``big model" (full model), whereas a ``little model" can be defined as one that activates fewer than \(K\) experts, which is sufficient for processing most tokens. We set the number of activated experts for the ``little model" to \(\frac{K}{2}\) for MoBiLE.

Traditional speculative decoding methods reduce the overhead associated with autoregressive decoding. In the case of MoE inference, the significant overhead arises from the ineffective prefetching function and the considerable time spent loading experts to GPU. 
By utilizing the little model, the number of experts that need to be loaded is reduced, resulting in a substantial decrease in loading time overhead, as shown in Fig. \ref{fig3}.
For the big model, we directly employ the default number of experts selected by the little model for inference, effectively treating the process of generating the evicted token by the little model as a prefetching function.

\begin{algorithm}
\caption{MoBiLE inference algorithm}
\label{alg2}
\begin{algorithmic}[1]
\State {\bfseries Input:} $input\_tokens$
\State $y \gets \langle input\_tokens \rangle$ 
\While{$y[-1] \neq \langle eos \rangle$}
    \State $p_s, h_s\gets \text{LittleModel}(y)$ \Comment{$h_s: \text{router states}$}
    \If{$\max(p_s[-1]) > \gamma$}
        \State $y \gets y + [\text{sample}(p_s[-1])]$
    \Else
        \State $p_l \gets \text{BigModel}(y,h_s)$
        \State $y \gets y + [\text{sample}(p_l[-1])]$

    \EndIf
\EndWhile
\State
\Return $y$
\end{algorithmic}
\end{algorithm}

The process of MoBiLE inference is illustrated in Algorithm \ref{alg2}.
For the little model, fewer experts are activated. After processing all decoder layers, we compare the highest score of the output logits with a threshold \(\gamma\). If it's lower than \(\gamma\), the current token is evicted and fallback is needed. We use the big model to regenerate the token, prefetching is used in this process.
If no fallback is required, we use the little model to continue generating the next token.

\textbf{How is MoBiLE different from existing speculative decoding strategies?} 
Existing speculative decoding methods \cite{DBLP:conf/nips/KimMMMMGK23} typically require both a little model and a large model, making the selection of suitable models a critical challenge. Training little models poses difficulties in achieving a balance across various domains and necessitates considerable computational resources. In contrast, MoBiLE derives the little model from the large model, thereby retaining more of the large model's capabilities without requiring additional training. MoBiLE also eliminates the need for extra storage and the extra KV cache of the little model.

Moreover, traditional speculative decoding methods \cite{DBLP:conf/nips/KimMMMMGK23} discard the token generated by the little model if a fallback is necessary, leading to a waste of the computational effort involved in generating the evicted token. MoBiLE optimizes this process by leveraging the evicted token for prefetching the experts required by the large model, effectively addressing the challenges associated with expert prefetching in MoE models. 

\begin{table*}[t]
\begin{center}
\begin{threeparttable}
\caption{Accuracy results of Qwen MoE and OLMoE with different methods on GSM8K and Humaneval.}
\label{table3}
\begin{tabular}{lcccccccccc}
\toprule
\multirow{2}{*}{\textbf{Model}} & \multirow{2}{*}{\textbf{Method}} & \multicolumn{3}{c}{\textbf{GSM8K}} & \multicolumn{3}{c}{\textbf{Humaneval}} \\
\cmidrule(lr){3-5} \cmidrule(lr){6-8}
& & Accuracy & Fallback ratio & Speedup & Pass@1 & Fallback ratio & Speedup \\

\midrule
\multirow{3}{*}{Qwen MoE} 
& 4/64* (vanilla) & 0.641 & - & 1.00\(\times\) & 0.262 & - & 1.00\(\times\) \\ 
& 2/64** (vanilla) & 0.556 & 0.0 & 1.82\(\times\) & 0.146 & 0.0 & 1.82\(\times\) \\ 
& \textbf{2/64 + MoBiLE} & \textbf{0.608 (\textcolor{red}{+ 0.052})} & 0.11 & 1.72\(\times\) & \textbf{0.256  (\textcolor{red}{+ 0.110})} & 0.11 & 1.72\(\times\)  \\

\midrule
\multirow{3}{*}{OLMoE} 
& 8/64 (vanilla) & 0.515 & - & 1.00\(\times\) & 0.116 & - & 1.00\(\times\) \\ 
& 4/64  (vanilla)& 0.371 & 0.0 & 1.85\(\times\) & 0.098 & 0.0 & 1.85\(\times\) \\ 
& \textbf{4/64 + MoBiLE} & \textbf{0.448 (\textcolor{red}{+ 0.077})} & 0.21 & 1.57\(\times\) & \textbf{0.116 (\textcolor{red}{+ 0.018})} & 0.12 & 1.63\(\times\) \\
\bottomrule
\end{tabular}
\begin{tablenotes}
\small
\item[*] This is the default setting of the MoE model, which activates 4 out of 64 experts.
\item[**] In this setting, we reduce the number of activated experts by half, which only activates 2 out of 64 experts.
\end{tablenotes}
\end{threeparttable}
\end{center}
\end{table*}

\subsection{Fallback and Prefetch Function} \label{FPF}
In this section, we illustrate the fallback and prefetch functions in detail. As shown in Fig. \ref{fig3}, if we only activate fewer experts, the low latency comes with low accuracy, which limits the applicability of the model. The fallback function is aimed at preserving the accuracy performance of the model. 

\begin{figure}[htbp]
    \centering
    \includegraphics[width=0.85\columnwidth,keepaspectratio]{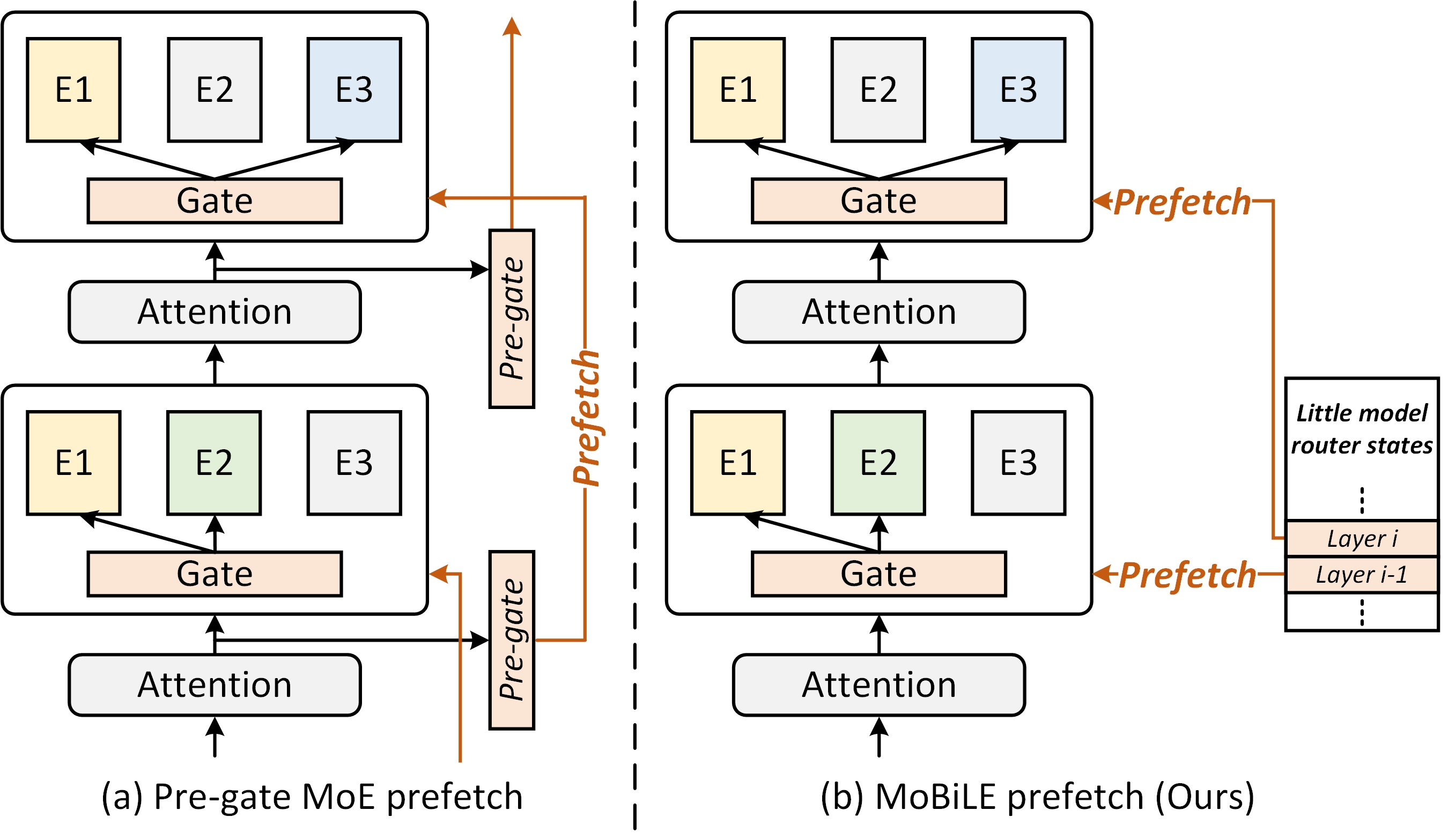} 
    \caption{Comparison of pre-gated MoE prefetch and MoBiLE prefech.}
    \label{fig4}
\end{figure}

When the little model generates a token, it chooses the highest score from the final output logits. In certain cases, the score may be very high (close to 1), indicating strong confidence in the correctness of the output. Conversely, low scores suggest that the little model may struggle to effectively process the current token, necessitating a fallback to the large model for accurate generation.

Although the token generated by the little model is evicted, we can still leverage the information from this process. The expert selection scores of the little model can be effectively used to prefetch the experts for the big model. 
For example, the default number of experts activated is 4, we activate top-2 experts for the little model. When the little model doesn't generate confident results and fallback is needed, the newly generated token is evicted. We perform prefetching for the big model \textbf{based on the router outputs from the previous evicted token}. 
This prefetching is different from existing pre-gating based prefetching methods, as shown in Fig. \ref{fig4}. We can identify the target experts of all the layers in the beginning, and we can dynamically adjust the prefetching time. For hardware configurations with low versions of PCIe connections, we may need to prefetch the experts 2 or 3 layers ahead to fully overlap the expert loading latency. This is difficult to achieve for pre-gated MoE because the gating function is difficult to give accurate predictions 2 layers ahead. Additionally, our prefetching is a plug-and-play method, leading to no additional training cost.

Fig. \ref{fig3} illustrates the workflow of MoBiLE big model and little model. The little model achieves speedup compared with the standard method due to the fewer loaded and executed experts. The big model uses prefetching to fully overlap the expert loading time, thus giving further speedup compared with the standard method.

Suppose the latency for full MoE offloading, MoBiLE little expert, and MoBiLE big expert to generate a token is \(T\), \(T_l\) and \(T_b\), respectively. The fallback ratio is \(r\). The speedup can be estimated by the following equation.

\begin{equation}
speedup = \frac{T}{T_l+r\times T_b} = \frac{1}{\frac{T_l}{T} + r \times \frac{T_b}{T}}
\end{equation}

\textbf{How is MoBiLE different from existing offloading-based MoE acceleration methods?}
Existing MoE prefetch methods need to train the prefetch function, which usually results in poor accuracy performance and requires much computational effort, as mentioned in Section \ref{Background and Motivation}. MoBiLE doesn't require any training, and based on our experiments, the accuracy of using the experts selected by the little model can achieve an accuracy of more than 80\%, the fallback with prefetching can gain accuracy increase as shown in Section \ref{Evaluation}. What's more, existing MoE prefetch methods mainly focus on Google Switch Transformers \cite{DBLP:journals/jmlr/FedusZS22}, which only activate 1 expert for each MoE module. MoBiLE pioneers effective inference for the newly proposed MoE architecture with finer-grained expert partition.

MoBiLE can also be \textbf{combined with existing prefetching methods} to gain further speedup. For the execution of the little model, we can utilize existing prefetching methods and train prefetching gates for prefetching. With prefetching, the little model can gain further speedup.

\section{Evaluation}  \label{Evaluation}

We conduct our evaluation under the following settings. We use 2 recent SOTA MoE models, including OLMoE \cite{DBLP:journals/corr/abs-2409-02060} and Qwen 1.5 MoE \cite{qwen_moe} for experiments. OLMoE activates 8 out of 64 experts, and Qwen MoE activates 4 out of 60 experts.
We test the accuracy results of these MoE models with different settings on generative tasks, including 0-shot Humaneval \cite{DBLP:journals/corr/abs-2107-03374} and 8-shot GSM8K \cite{DBLP:journals/corr/abs-2110-14168}. 
\begin{itemize}
    \item \textbf{GSM8K \cite{DBLP:journals/corr/abs-2110-14168}} contains math problems for which the model needs to generate the answering process and final answer. The result of GSM8K can show the model's ability in reasoning. 
    \item \textbf{Humaneval \cite{DBLP:journals/corr/abs-2107-03374}} consists of a set of Python programming problems, each accompanied by a prompt and a set of unit tests. The benchmark evaluates models' programming ability.
\end{itemize}

For these models and tasks, we set the fallback threshold \(\gamma\) to 0.7. The max generation length for GSM8K and Humaneval is set to 256 and 512, respectively.
Further ablation studies on the fallback threshold selection are in Section \ref{ablation}.

Then we evaluate the performance of MoBiLE models with different fallback ratios on a consumer hardware system containing a RTX 4080 GPU with 16GB HBM, an Intel i5-12400 CPU with 64GB DRAM, the connection between CPU and GPU is PCIe 4.0. The setting is a usual configuration of consumer laptops.
All 2 models can run on the hardware platform with offloading. We evaluate the end-to-end speedup of different models and methods. 
 
\subsection{Accuracy Results}
As shown in Table \ref{table3}, directly reducing the number of activated experts leads to a significant drop in accuracy. For Qwen MoE, it leads to a \(8.5\%\) and a \(11.6\%\) decrease in accuracy on GSM8K and Humaneval, respectively. For OLMoE, it leads to a \(14.4\%\) and \(1.8\%\) decrease in accuracy on GSM8K and Humaneval, respectively. 
Although they show great speedup, this severe accuracy degradation limits the actual usage of these vanilla methods.

When using MoBiLE for inference, the accuracy of the tasks increases compared with naively activating fewer experts. Qwen MoE MoBiLE achieves an accuracy of \(60.8\%\) and \(25.6\%\) on GSM8K and Humaneval, and OLMoE MoBiLE achieves an accuracy of \(44.8\%\) and \(11.6\%\) on GSM8K and Humaneval. 
The accuracy results of MoBiLE significantly surpass the vanilla method, indicating the effectiveness of the fallback mechanism. The low accuracy degradation makes MoBiLE promising in actual scenarios of generation tasks.

\subsection{System Performance}
We measure the end-to-end speedup of the 2 MoE models on the 2 evaluation tasks compared with the default offload setting. 
The results are shown in Table \ref{table3}. MoBiLE achieves an average  speedup of \(1.72\times\) and \(1.60\times\) for Qwen MoE and OLMoE compared with the full model baseline.

\begin{figure}[htbp]
    \centering
    \includegraphics[width=0.9\columnwidth,keepaspectratio]{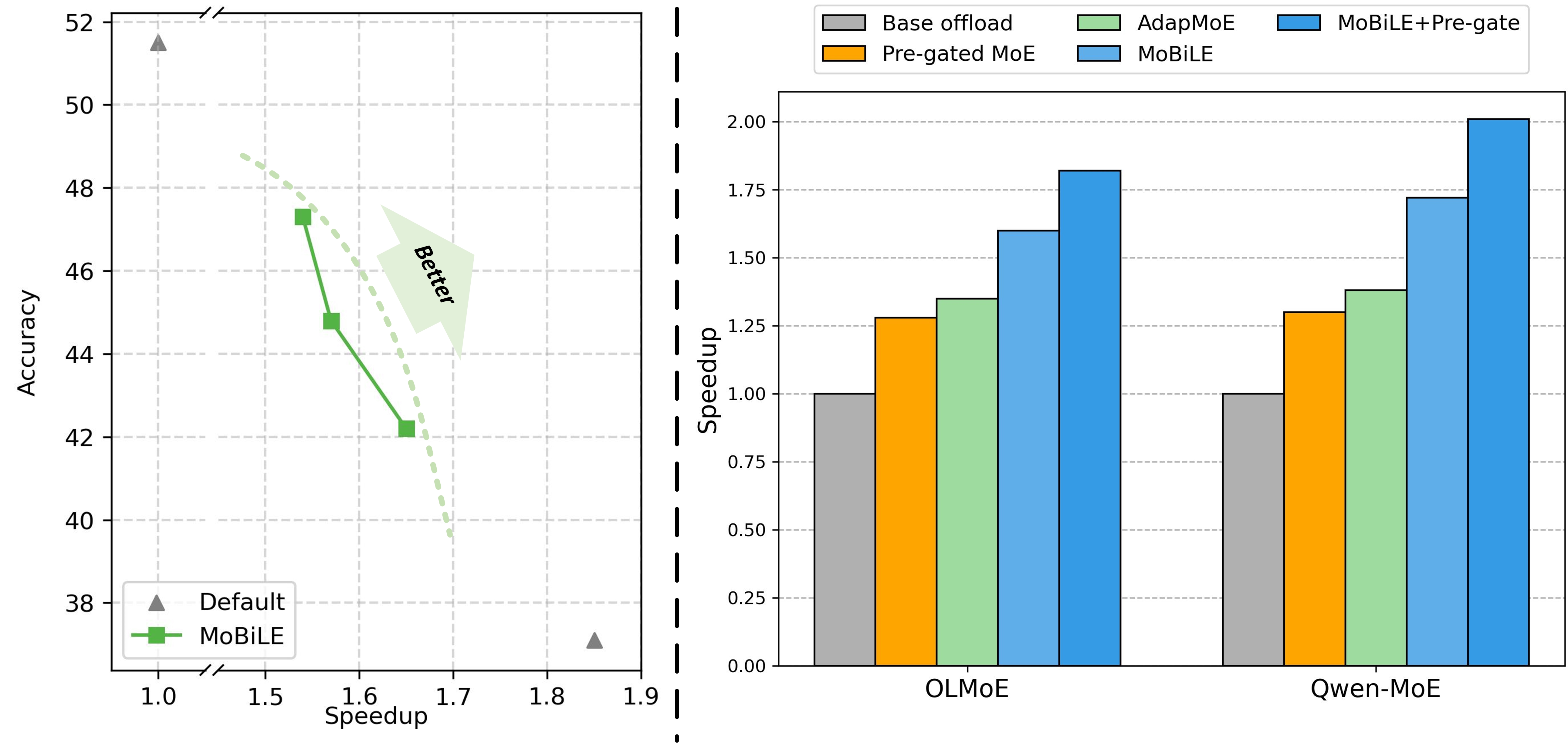} 
    \caption{Left: Accuracy-efficiency tradeoff for MoBiLE. Right: Speedup of MoBiLE compared with SOTA MoE offloading methods.}
    \label{speedup}
\end{figure}

We further compare the speedup of MoBiLE with SOTA MoE offloading methods Pre-gated MoE \cite{DBLP:conf/isca/HwangWCHTCY24} and AdapMoE \cite{DBLP:journals/corr/abs-2408-10284}. To ensure a fair comparison, all methods are evaluated under an identical constraint for maximum GPU memory utilization.
The results are shown in Fig. \ref{speedup}. MoBiLE achieves an average speedup of \(1.28\times\) and \(1.21\times\) compared with Pre-gated MoE and AdapMoE, respectively. 

Furthermore, we assessed an integrated configuration where MoBiLE is combined with pre-gating modules, as described in Section \ref{FPF}. 
It achieves an average speedup of \(1.92\times\) compared with the offloading baseline, significantly outperforming existing MoE offloading methods which require additional training for the prediction modules.

\section{Ablation Study} \label{ablation}
In this section we conduct ablation studies for MoBiLE, including the accuracy-efficiency tradeoff and larger design space exploration. 

\subsection{Accuracy-efficiency Tradeoff}
We conduct further experiments on OLMoE and GSM8K task to show the accuracy-efficiency tradeoff of MoBiLE system. We set the fallback threshold to 0.5, 0.7, and 0.8. The fallback ratio and accuracy are shown in Table. \ref{table4}.

As the threshold increases, the fallback ratio increases, resulting in higher accuracy. However, as the fallback ratio increases, the latency also increases, thus leading to poorer speedup. 
\textbf{MoBiLE doesn't need to tune \(\gamma\) for different models and tasks}. The default setting of \(\gamma = 0.7\) achieves a balance between accuracy and speedup, offering superior speedup without compromising accuracy performance, as shown in Fig. \ref{speedup}. 

\subsection{Design Space Exploration}
We conduct an extensive study on activating fewer or more experts for the little model on OLMoE and GSM8K task. We choose to activate 2 or 6 experts, and the fallback threshold \(\gamma\) is set to 0.7.

The results are shown in Table. \ref{table5}. The MoE model activating fewer than half of the experts (2/64) shows a significant poor accuracy performance of \(4\%\). After applying MoBiLE, the performance increased to \(18\%\), indicating the effectiveness of the fallback and prefetching function. 
When activating more than half of the experts (6/64), the model shows improved accuracy; nonetheless, the speedup is constrained due to the increased computational and memory overhead associated with the smaller model.
These results indicate that our setting of activating half of the default number of experts to serve as the little model for MoBiLE achieves superior performance in accuracy and speedup.

\begin{table}[t]
\begin{center}
\caption{Ablation study on the performance of MoBiLE with different fallback thresholds with OLMoE on GSM8K task.}
\label{table4}
\begin{tabular}{cccc}
\toprule
\textbf{Threshold} & \textbf{Accuracy} & \textbf{Fallback ratio} & \textbf{Speedup}  \\
\midrule
\(\gamma\) = 0.0 & 0.371 & 0.0 & 1.85\(\times\)\\
\(\gamma\) = 0.5 & 0.422 & 0.10 & 1.65\(\times\)\\
\(\gamma\) = 0.7 & 0.448 & 0.21 & 1.57\(\times\)\\
\(\gamma\) = 0.8 & 0.473 & 0.27 & 1.54\(\times\)\\
\bottomrule
\end{tabular}

\end{center}
\end{table}

\begin{table}[!t]
\begin{center}
\caption{Ablation study on activating fewer experts for MoBiLE with OLMoE on GSM8K task.}
\label{table5}
\begin{tabular}{lccc}
\toprule
\textbf{Method} & \textbf{Accuracy} & \textbf{Fallback ratio}  \\
\midrule
8/64 (vanilla) & 0.515 & - \\
\midrule
2/64 (vanilla) & 0.040 & 0.0 \\
\textbf{2/64 + MoBiLE} & 0.180 & 0.29 \\
4/64 (vanilla) & 0.371 & 0.0 \\
\textbf{4/64 + MoBiLE} & 0.448 & 0.21 \\
6/64 (vanilla) & 0.491 & 0.0 \\
\textbf{6/64 + MoBiLE} & 0.506 & 0.21 \\
\bottomrule
\end{tabular}

\end{center}
\end{table}

\section{Conclusion}
In this paper we propose MoBiLE, an efficient offloading-based inference system for MoE models. MoBiLE is, as far as we know, the first efficient inference design for up-to-date MoE architectures. It is also the first work enabling MoE inference on storage-limited consumer-level hardware. 
By activating fewer experts for most tokens and performing fallback with efficient prefetching for important tokens, MoBiLE not only achieves an inference speedup of \(1.60\times\) to \(1.72\times\) compared with offloading baseline on consumer hardware, but it also leads to negligible degradation in accuracy compared with the full model.

\section*{Acknowledgment}
This work was supported in part by the NSFC under Grant 62304121,  Grant 62125403, and Grant 92164301; in part by the National Key R\&D Project under Grant 2023YFB4403100; in part by the National Science and Technology Major Project under Grant 2022ZD0115201; in part by the National Key Research and Development Program under Grant 2021ZD0114400; Beijing S\&T Project Z221100007722023; in part by Beijing National Research Center for Information Science and Technology; and in part by Beijing Advanced Innovation Center for Integrated Circuits. 

\bibliographystyle{IEEEtranS}
\bibliography{main}
\end{document}